\documentclass{article}

\usepackage{arxiv}
\usepackage{eccvabbrv}
\usepackage{multirow}
\usepackage{graphicx}
\usepackage{booktabs}
\usepackage{amsmath,amssymb}
\usepackage{mathtools}
\usepackage{bm}
\usepackage{bbm}
\usepackage{subcaption}
\usepackage{xcolor}
\definecolor{eccvblue}{RGB}{0,102,204}
\usepackage{mathrsfs}   
\usepackage{microtype}
\usepackage[table]{xcolor}
\definecolor{surgblue}{RGB}{235,243,255}   
\definecolor{surgpurple}{RGB}{243,236,255} 
\usepackage[accsupp]{axessibility}  
\usepackage{booktabs} %
\usepackage{tabularx} %
\usepackage{array} %
\usepackage{makecell}
\usepackage{caption}
\usepackage{siunitx}
\usepackage[pagebackref,breaklinks,colorlinks,citecolor=eccvblue]{hyperref}
\usepackage{orcidlink}
\begin{document}

\title{\textbf{SurgFormer: Scalable Learning of Organ Deformation with Resection Support and Real-Time Inference}}

\author{
{\normalfont Ashkan Shahbazi}$^{1,4}$\thanks{Equal contribution.}
\and
{\normalfont Elaheh Akbari}$^{1,4}$\footnotemark[1]
\and
{\normalfont Kyvia Pereira}$^{2,4}$
\and
{\normalfont Jon S.~Heiselman}$^{2,4}$
\and
{\normalfont Annie Benson}$^{2,4}$
\and
{\normalfont Garrison L.~H.~Johnston}$^{3,4}$
\and
{\normalfont Jie Ying Wu}$^{1,4}$
\and
{\normalfont Nabil Simaan}$^{3,4}$
\and
{\normalfont Michael I.~Miga}$^{2,4}$
\and
{\normalfont Soheil Kolouri}$^{1,4}$
\\[1em]
$^1$Department of Computer Science, College of Connected Computing, Vanderbilt University
\\
$^2$Department of Biomedical Engineering, Vanderbilt University
\\
$^3$Department of Mechanical Engineering, Vanderbilt University
\\
$^4$Vanderbilt Institute for Surgery and Engineering
}

\maketitle

\begin{abstract}
We introduce SurgFormer, a multiresolution gated transformer for data driven soft tissue simulation on volumetric meshes. High fidelity biomechanical solvers are often too costly for interactive use, so we train SurgFormer on solver generated data to predict nodewise displacement fields at near real time rates. SurgFormer builds a fixed mesh hierarchy and applies repeated multibranch blocks that combine local message passing, coarse global self attention, and pointwise feedforward updates, fused by learned per node, per channel gates to adaptively integrate local and long range information while remaining scalable on large meshes. For cut conditioned simulation, resection information is encoded as a learned cut embedding and provided as an additional input, enabling a unified model for both standard deformation prediction and topology altering cases. We also introduce two surgical simulation datasets generated under a unified protocol with XFEM based supervision: a cholecystectomy resection dataset and an appendectomy manipulation and resection dataset with cut and uncut cases. To our knowledge, this is the first learned volumetric surrogate setting to study XFEM supervised cut conditioned deformation within the same volumetric pipeline as standard deformation prediction. Across diverse baselines, SurgFormer achieves strong accuracy with favorable efficiency, making it a practical backbone for both tasks. {Code, data, and project page: \href{https://mint-vu.github.io/SurgFormer/}{available here}}. \textbf{Keywords:} Soft Tissue Deformation, Geometric Deep Learning, Surgical Simulation.
\end{abstract}

\begin{figure}[t]
  \centering
  \includegraphics[width=\textwidth]{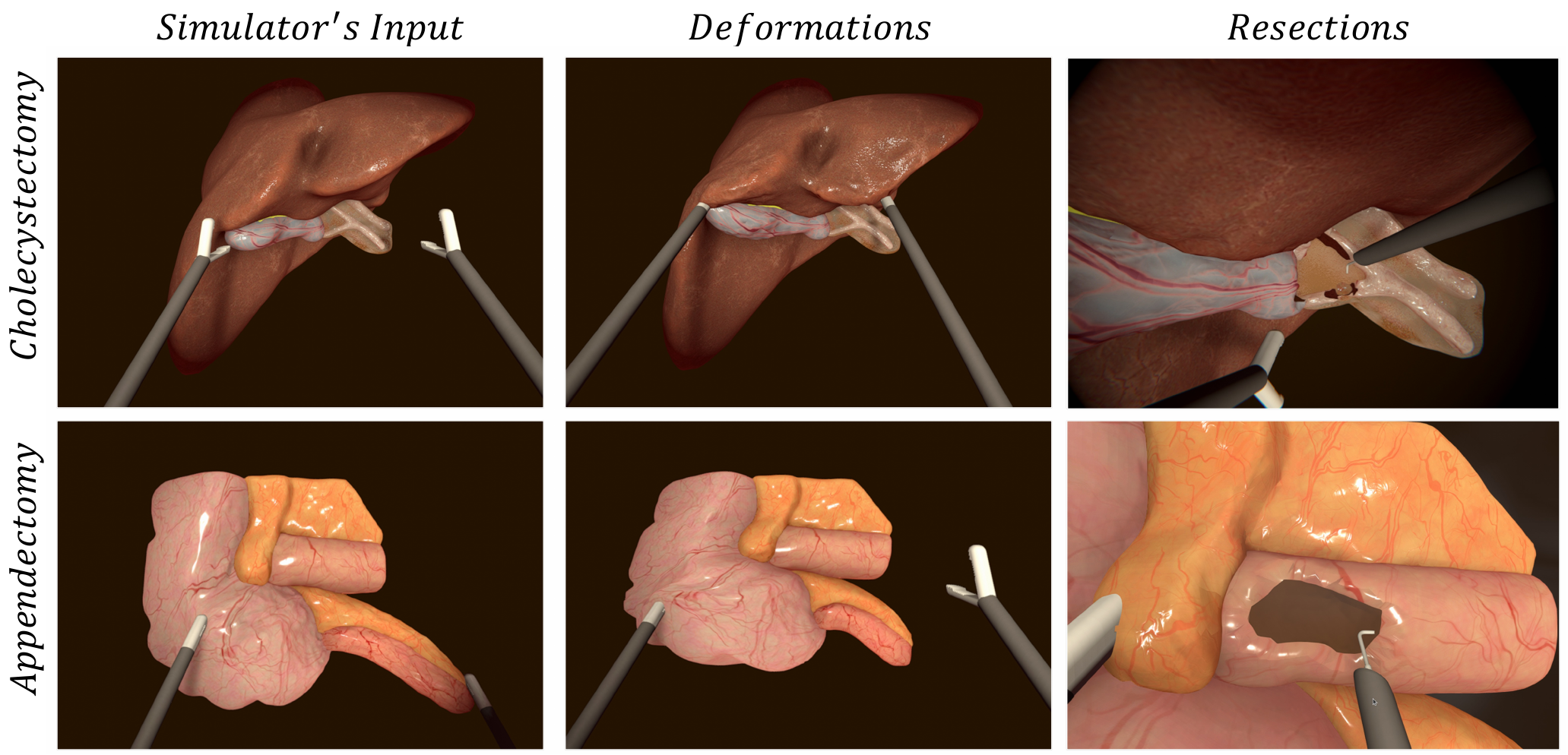}

  \caption{SurgFormer enables real-time, anatomically plausible soft-tissue simulation for minimally invasive surgery, jointly modeling tool-driven deformation and topology-changing resection. We demonstrate cholecystectomy (top) and appendectomy (bottom); each row shows the input, predicted deformation under tool interaction, and resulting resection state, highlighting geometric fidelity and visual realism.}
  \label{fig:teaser}
  \vspace{-0.2in}
\end{figure}

\section{Introduction}

Accurate modeling of soft tissue deformation is a core component of surgical training, planning, and deformation-aware guidance systems \cite{Han_2024,kojanazarova2025softtissuesimulationforce}. High-fidelity finite element methods remain the standard for volumetric biomechanics, but interactive use is limited by the cost of repeatedly solving large, sparse systems under contact and changing boundary conditions \cite{kojanazarova2025softtissuesimulationforce}. This has motivated learned surrogates to approximate full-field displacement responses from simulation data while targeting near-real-time inference \cite{pfaff2021learningmeshbasedsimulationgraph,shahbazi2026neuralaugmentedkelvinletrealtimesoft}.

Learning on surgical geometry differs from grid based PDE benchmarks because organs are naturally represented as irregular volumetric meshes with large node counts and heterogeneous element quality. Surgical interaction signals are sparse and localized, yet their effects can propagate globally through tissue, and cutting introduces discontinuities and changes in the active domain that are difficult to capture with purely local processing. In physics simulation, these discontinuities are naturally handled by XFEM style enrichment and element removal \cite{ABDELAZIZ20081141}, and recent graphics work continues to improve robust XFEM formulations for deformable cutting \cite{tonthat2024xfemcutting}.

Recent learned surrogate modeling spans graph based mesh simulators, transformer style global processors, and neural operator frameworks. Mesh based graph networks are effective at encoding local interactions and are strong baselines for learned simulation on irregular discretizations \cite{pfaff2021learningmeshbasedsimulationgraph}, while more recent hybrids add transformer style global interaction to improve long range information flow on large meshes \cite{iparraguirre2026mgntransformer}. In parallel, neural operators aim to learn solution operators with improved transfer across discretizations, including geometry informed operator learning on complex 3D domains \cite{li2023gino}, transformer based operator learners \cite{hao2023gnot}, and multiscale graph operator frameworks \cite{li2024amg}. However, applying dense global interaction at the finest mesh resolution remains costly for large volumetric organ meshes, especially in procedure level settings that require interactive latency. Moreover, most learned surrogates for surgical mechanics focus on continuous deformation and do not explicitly address cut conditioned deformation with topology changes within a unified volumetric learning pipeline.\cite{JMLR:v24:23-0064,ZHUANG20255416}

We introduce SurgFormer, a multiresolution gated transformer for global organ manipulation on volumetric meshes with near real time inference. SurgFormer builds a fixed mesh hierarchy and applies repeated multibranch blocks that combine local message passing, global self attention restricted to coarse levels, and pointwise updates fused by learned per node per channel gates. This design preserves efficiency at high resolution while enabling global context where it is affordable. In addition, to our knowledge, SurgFormer is the first learned volumetric surrogate in this setting to support both standard deformation prediction and cut conditioned deformation prediction within a single unified architecture and evaluation pipeline. Cut conditioning is achieved by augmenting each node's input feature vector — which includes geometric and mechanical quantities such as position and displacement — with a learned d-dimensional embedding that indicates whether the node belongs to the resected or intact region. The mesh graph remains fixed throughout, and the network learns to interpret these per-node cut condition embeddings jointly with all other node features to predict the resulting deformation under topology-changing surgical mechanics, with supervision from XFEM generated simulations.

\paragraph{Contributions.}
\begin{itemize}
\item We introduce SurgFormer, a multiresolution gated transformer for volumetric soft-tissue deformation prediction that combines local message passing, coarse-level global attention, and pointwise updates in a scalable unified block.
\item We formulate resection-aware cut-conditioned deformation prediction within the same FEM/XFEM supervision pipeline as standard deformation prediction, enabling a near-real-time learned volumetric surrogate for topology-changing surgical mechanics.
\item We introduce two procedure-level surgical simulation datasets under a unified protocol (cholecystectomy resection and appendectomy manipulation/resection with cut and uncut cases).
\item We provide unified accuracy/efficiency benchmarks, architectural ablations, transfer studies, and adversarial smoothness stress tests, showing strong performance with near-real-time inference.
\end{itemize}

\vspace{-0.1in}
\section{Related Work}
\label{sec:related_work}

\paragraph{Learning based surrogates for soft tissue mechanics.}
Fast surrogates for deformable mechanics increasingly complement full finite element solves when interactive rates are required. Graph based surrogates such as MGN\cite{pfaff2021learningmeshbasedsimulationgraph} learn mesh resolved dynamics from simulation data and are a common baseline for mesh simulation tasks. Recent work improves long range information flow on large meshes by augmenting mesh graph processing with transformer style global interaction, including MeshGraphNet-Transformer (MGN-T) \cite{iparraguirre2026mgntransformer}. Beyond purely data driven surrogates, several lines of work inject physical structure to improve generalization and stability, including graph neural networks informed locally by thermodynamics \cite{tierz2024localthermo} and Kelvinlet augmented learning that uses analytic deformation kernels as priors for real time soft tissue prediction \cite{shahbazi2026neuralaugmentedkelvinletrealtimesoft}.

\paragraph{Neural operators and attention on irregular geometries.}
Operator learning methods approximate solution operators rather than single instance mappings, enabling resolution transfer and improved sample efficiency. Geometry aware operator learning has advanced through methods that combine graph based processing on irregular discretizations with latent regular grid representations, such as GINO \cite{li2023gino}, and transformer based operator learners such as GNOT \cite{hao2023gnot}. More recent work extends this direction with multi graph operator frameworks that explicitly model scale and physics on arbitrary geometries, such as AMG \cite{li2024amg}, and geometry informed neural operator transformers for arbitrary geometries, such as GINOT \cite{liu2026ginot}. Complementary hybrid approaches accelerate classical nonlinear finite element solvers by using an operator learned predictor as an initializer followed by Newton refinement \cite{taghikhani2025nin}.

\paragraph{Cutting, discontinuities, and topology change.}
Modeling cutting and resection introduces discontinuities and changes in the active domain, which are naturally handled in physics simulation by XFEM style enrichment and topology modification \cite{ABDELAZIZ20081141}. Recent surgical simulation work continues to improve real time cutting realism through fracture criteria, rupture prediction, and efficient constraint based dynamics, including fracture mechanics plus PBD models for dura mater cutting \cite{shi2025dura_pbd_cutting}. While many learned deformable surrogates focus on continuous deformation, modeling cutting induced discontinuities and topology change is a key step toward realistic, real time tissue dynamics, since these events are central to procedural simulation yet violate the smooth deformation assumptions that many surrogates implicitly rely on. Accordingly, an emerging line of work aims to make learned models robust to sharp discontinuities and changing domains, for example by embedding stronger physics structure \cite{pfaff2021learningmeshbasedsimulationgraph,li2023gino}.

\noindent\paragraph{Positioning.}
SurgFormer targets learning based global tissue manipulation on large volumetric meshes, enabling near real time inference for both deformation prediction and cut conditioned modeling. To our knowledge, prior learned surgical mechanics surrogates have largely focused on continuous deformation, rather than a unified volumetric formulation that also supports cut conditioned prediction within the same architecture and protocol. This makes SurgFormer a practical backbone for interactive surgical training and procedure level evaluation, where low latency is essential.
\vspace{-0.1in}
\section{Method}
\label{sec:method}

\begin{figure*}[t]
  \centering
  \includegraphics[width=\textwidth]{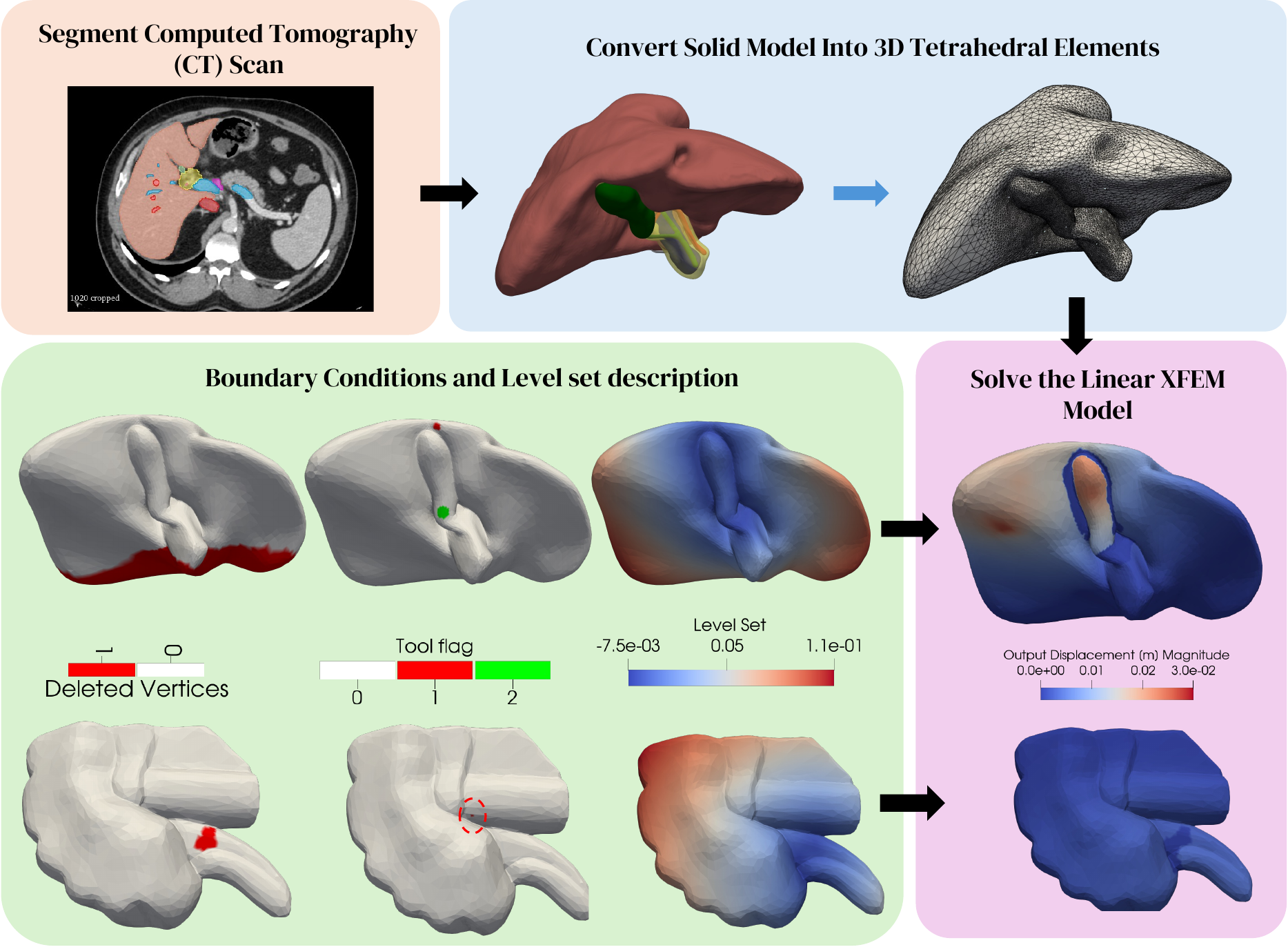}
  \caption{XFEM based data generation pipeline for surgical tissue mechanics. CT volumes are segmented to obtain organ geometries, converted to solid models, and tetrahedralized into volumetric meshes. Procedure-specific boundary conditions, tool interaction flags, deleted-vertex indicators, and level-set fields are then defined to represent deformation and resection states. Quasistatic linear XFEM is solved on the active domain to produce full-field displacement targets. \vspace{-0.24in}}
  \label{fig:fem}
\end{figure*}

\subsection{Biomechanics and data generation}
\label{sec:biomech}

We construct supervised pairs by solving quasistatic linear elasticity with XFEM on tetrahedral organ meshes, so that resection-induced displacement jumps are represented explicitly. Figure~\ref{fig:fem} summarizes the pipeline: organ geometries are segmented from CT in 3D Slicer~\cite{fedorov20123d}, converted to solids, and tetrahedralized with \texttt{gmsh}~\cite{geuzaine_gmsh_2009}. The resulting meshes contain \(7{,}870\) vertices and \(29{,}723\) tetrahedra for appendectomy, and \(13{,}085\) vertices and \(48{,}911\) tetrahedra for cholecystectomy. To model resection, we use an XFEM displacement field with a Heaviside enrichment across the cut interface:
\begin{equation}
u(\xi)=\sum_{i=1}^{m} N_i(\xi)\,\tilde u_i
+\sum_{j\in\mathcal I_{\mathrm{enr}}} N_j(\xi)\big(H(\xi)-H(\xi_j)\big)\,\tilde a_j,
\qquad \xi \in \Omega,
\end{equation}
where \(N_i\) are linear tetrahedral shape functions, \(\tilde u_i\) are standard nodal unknowns, \(\mathcal I_{\mathrm{enr}}\) is the set of enriched nodes, \(H\) is the discontinuous Heaviside function, and \(\tilde a_j\) are enrichment coefficients. The resection interface is encoded by a level set \(\phi(\xi)\): cut elements with \(\phi(\xi)=0\) are enriched, while elements in the removed region \(\phi(\xi)<0\) are excluded from the active domain. Detached vertices are additionally flagged and passed to the network as topology indicators. We solve linear elasticity on the active domain in \texttt{getFEM}~\cite{renard_getfem_2020} with \((E,\nu)=(2100\,\mathrm{Pa},0.45)\), producing full-field displacement solutions over intact and resection states, with \(120{,}000\) samples for cholecystectomy and \(320{,}000\) samples for appendectomy. For cholecystectomy, we simulate 25 progressive resections along the gallbladder-liver interface and impose Dirichlet constraints near the bare area of the liver and distal regions of the vena cava and portal vein. For appendectomy, we simulate 25 progressive resections separating the mesoappendix and appendix, with the superior cecal region constrained. The choice of fixed regions and procedure-specific resection progression is defined using prior anatomical and procedural knowledge from a medical expert. Tool actions are modeled as prescribed boundary displacements on sampled surface interaction vertices \(p_s\), excluding fixed regions. More precisely, we sample an interaction index \(s\in\mathcal V_{\mathrm{surf}}\) uniformly from admissible surface nodes and use its position \(p_s\). We draw displacement magnitudes from \(\mathcal U(-30,70)\,\mathrm{mm}\), and sample directions within a cone around the local outward normal with half-angle \(\pi/5\). This yields grasping, pulling, poking, and pushing-like interactions under a unified boundary actuation model.

\begin{figure*}[t]
  \centering
  \includegraphics[width=\textwidth]{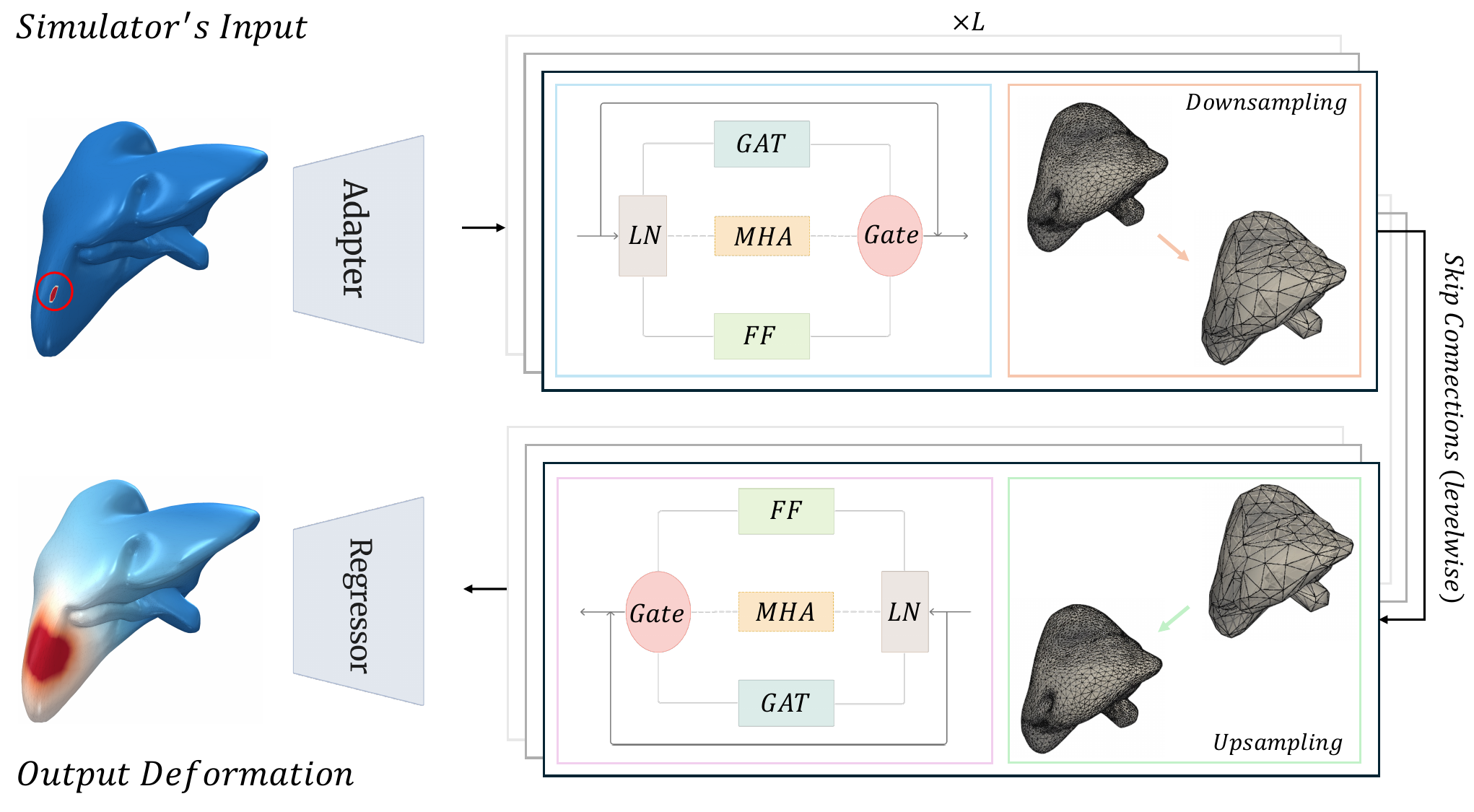}
  \caption{Overview of the proposed multi resolution graph transformer for organ deformation. Simulator input signals are encoded by an Adapter and processed through $L$ hierarchical levels using farthest point sampling with pooling, where each level applies LayerNorm followed by parallel local graph attention, global multi head attention, and feed forward transformations fused by a learned gate. A symmetric decoder upsamples via unpooling and refines features at each level, using levelwise skip connections from the encoder, and a final regressor maps the finest level features to the output. \vspace{-0.2in}}
  \label{fig:overview}
\end{figure*}

\vspace{-0.1in}
\subsection{Problem setup and notation}
\vspace{-0.02in}
We predict a nodewise displacement field on a volumetric mesh. Let the finest-level mesh graph be
\(G^{(0)}=(V^{(0)},E^{(0)})\) with \(|V^{(0)}|=N_0\). Each vertex \(i\in V^{(0)}\) has Euclidean position
\(p_i\in\mathbb{R}^3\) and a raw input feature vector \(f_i\in\mathbb{R}^{d_{\mathrm{in}}}\).
Stacking raw node features gives \(F^{(0)}\in\mathbb{R}^{N_0\times d_{\mathrm{in}}}\).
The target nodewise displacement field is \(U\in\mathbb{R}^{N_0\times 3}\), and the model prediction is
\(\hat U\in\mathbb{R}^{N_0\times 3}\). We build a fixed multiresolution hierarchy with levels \(\ell\in\{0,\dots,L\}\).
At level \(\ell\), the graph is \(G^{(\ell)}=(V^{(\ell)},E^{(\ell)})\) with \(|V^{(\ell)}|=N_\ell\) and hidden node features
\(X^{(\ell)}\in\mathbb{R}^{N_\ell\times D}\). Since \(V^{(\ell)}\subseteq V^{(0)}\) under our construction, each
vertex \(i\in V^{(\ell)}\) inherits the Euclidean position \(p_i\) from the original mesh. We use \(N_h\) attention heads
with head dimension \(d=D/N_h\).

\vspace{-0.1in}
\subsection{Fixed multiresolution hierarchy}
\label{sec:hierarchy}

\paragraph{Seed selection by farthest point sampling.}
For each transition $\ell \to \ell+1$, we select a seed set $S^{(\ell+1)} \subseteq V^{(\ell)}$ with
$|S^{(\ell+1)}| = N_{\ell+1}$. We use farthest point sampling in Euclidean space using vertex positions $\{p_i\}$.
Initialize $S=\{s_1\}$ with a uniformly sampled vertex $s_1 \in V^{(\ell)}$ and maintain
$d_S(i) = \min_{s \in S} \|p_i - p_s\|_2$. Repeat until $|S| = N_{\ell+1}$:
\begin{equation}
s_{t+1} = \arg\max_{i \in V^{(\ell)}} d_S(i).
\end{equation}
After adding $s_{t+1}$, update $d_S(i) \leftarrow \min(d_S(i), \|p_i-p_{s_{t+1}}\|_2)$ for all $i \in V^{(\ell)}$.
We set $S^{(\ell+1)} := S$ and define $V^{(\ell+1)} := S^{(\ell+1)}$.

\paragraph{Ownership map and clusters.}
Each fine node is assigned to its nearest seed, defining an ownership map
$o^{(\ell+1)}: V^{(\ell)} \to V^{(\ell+1)}$:
\begin{equation}
o^{(\ell+1)}(i) = \arg\min_{s \in V^{(\ell+1)}} \mathrm{dist}_{G^{(\ell)}}(i,s),
\vspace{-0.1in}
\end{equation}
where $\mathrm{dist}_{G^{(\ell)}}$ is shortest path distance on $G^{(\ell)}$.
This induces clusters for each coarse node $s \in V^{(\ell+1)}$:
\begin{equation}
C_s^{(\ell+1)} = \{ i \in V^{(\ell)} : o^{(\ell+1)}(i) = s \}.
\vspace{-0.1in}
\end{equation}

\paragraph{Coarse edges.}
We induce coarse edges by contracting fine edges under $o^{(\ell+1)}$, that is
$(s,t)\in E^{(\ell+1)}$ if there exists $(u,v)\in E^{(\ell)}$ with
$o^{(\ell+1)}(u)=s$, $o^{(\ell+1)}(v)=t$, and $s\neq t$.
All hierarchy objects are computed once and kept fixed during training and inference.

\paragraph{Downsampling by channelwise max pooling.}
Given fine features $X^{(\ell)}$, we pool to coarse features $X^{(\ell+1)}$ using channelwise max pooling over clusters:
\begin{equation}
X^{(\ell+1)}_{s,c} = \max_{i \in C_s^{(\ell+1)}} X^{(\ell)}_{i,c},
\qquad
s \in V^{(\ell+1)},\; c \in \{1,\dots,D\}.
\label{eq:maxpool}
\vspace{-0.1in}
\end{equation}

\paragraph{Upsampling by broadcast.}
Given coarse features $Y^{(\ell+1)} \in \mathbb{R}^{N_{\ell+1}\times D}$, we upsample to the fine level by broadcast:
\begin{equation}
U_\ell(Y^{(\ell+1)})_{i,:} = Y^{(\ell+1)}_{o^{(\ell+1)}(i),:},
\qquad
i \in V^{(\ell)}.
\label{eq:broadcast}
\vspace{-0.1in}
\end{equation}

\vspace{-0.1in}
\subsection{SurgFormer block}
\label{sec:block}

Each level uses a multibranch block that mixes local message passing (i.e., local token mixing), optional global self attention (i.e., global token mixing), and a feedforward
update (i.e., feature mixing) via learned per-node per-channel gates. We use a pre-norm residual structure. Let
$\bar X^{(\ell)} := \mathrm{LN}(X^{(\ell)})$ denote layer normalized features.

\paragraph{Local message passing branch.}\label{sec:local_branch}
We use a GAT-style local attention layer on the directed edge list \(E^{(\ell)}\).
Let \(W^{(\ell)}\in\mathbb{R}^{D\times (N_h d)}\), and let
\(\psi^{(\ell),h}\in\mathbb{R}^{N_\ell\times d}\) denote the \(h\)-th head slice of
\(\bar X^{(\ell)}W^{(\ell)}\). For each head \(h\), we learn
\(a_{\mathrm{src}}^{(\ell),h},a_{\mathrm{dst}}^{(\ell),h}\in\mathbb{R}^{d}\). For an edge \((j,i)\in E^{(\ell)}\),
\begin{equation}
e_{ji}^{(\ell),h}
=\sigma\!\Big(
\langle a_{\mathrm{src}}^{(\ell),h},\psi_{j}^{(\ell),h}\rangle
+\langle a_{\mathrm{dst}}^{(\ell),h},\psi_{i}^{(\ell),h}\rangle
\Big),
\qquad
\alpha_{ji}^{(\ell),h}
=\mathrm{Softmax}_{j:(j,i)\in E^{(\ell)}}\!\big(e_{ji}^{(\ell),h}\big),
\end{equation}
\begin{equation}
O_{i}^{(\ell),h}
=\sum_{j:(j,i)\in E^{(\ell)}} \alpha_{ji}^{(\ell),h}\,\psi_{j}^{(\ell),h}.
\end{equation}
Concatenating heads gives
\begin{equation}
\Delta_{\mathrm{loc}}^{(\ell)}=\mathrm{Concat}\!\big(O^{(\ell),1},\dots,O^{(\ell),N_h}\big)\in\mathbb{R}^{N_\ell\times D}.
\end{equation}

\paragraph{Global self attention branch.} \label{sec:global_branch}
The global branch applies dense multihead self attention to \(\bar X^{(\ell)}\).
Define \(Q^{(\ell)}=\bar X^{(\ell)}W_Q^{(\ell)},\; K^{(\ell)}=\bar X^{(\ell)}W_K^{(\ell)},\; V^{(\ell)}=\bar X^{(\ell)}W_V^{(\ell)}\), with head slices \(Q^{(\ell),h},K^{(\ell),h},V^{(\ell),h}\in\mathbb{R}^{N_\ell\times d}\), we compute
\begin{equation}
A^{(\ell),h}=\mathrm{Softmax}\!\left(\frac{Q^{(\ell),h}(K^{(\ell),h})^\top}{\sqrt d}\right),
\qquad
Y^{(\ell),h}=A^{(\ell),h}V^{(\ell),h}
\end{equation}
Concatenating heads and projecting gives
\begin{equation}
\Delta_{\mathrm{glob}}^{(\ell)}=\mathrm{Concat}(Y^{(\ell),1},\dots,Y^{(\ell),N_h})W_O^{\mathrm{glob}},
\qquad
W_O^{\mathrm{glob}}\in\mathbb{R}^{D\times D}
\end{equation}
We include the dense global branch only at coarser hierarchy levels to reduce computational and memory cost while preserving long range interactions where they are most beneficial. For additional efficiency and memory savings, we implement the global attention computation with FlashAttention \cite{dao2022flashattention}, which yields the same scaled dot product attention output as standard attention but uses an optimized fused kernel.

\paragraph{Feedforward branch.}\label{sec:ff_branch} The feedforward branch is a pointwise MLP applied to each node:
\begin{equation}
\Delta_{\mathrm{ff}}^{(\ell)}=ReLU\!\left(\bar X^{(\ell)}W_1+\mathbf{1}{a_1}^\top\right)W_2+\mathbf{1}{a_2}^\top,
\qquad
W_1\in\mathbb{R}^{D\times D_{\mathrm{hid}}},\;
W_2\in\mathbb{R}^{D_{\mathrm{hid}}\times D}.
\end{equation}
Here \(a_1\in\mathbb{R}^{D_{\mathrm{hid}}}\) and \(a_2\in\mathbb{R}^{D}\) are biases.

\paragraph{Gating and fusion mechanism.}\label{sec:gating} Let $\mathcal{B}_\ell$ be an ordered list of active branches at level $\ell$, and let
$\Delta^{(\ell,b)}$ denote the corresponding proposal, for example local, global, or feedforward.
We compute gate logits and normalize across the $B_\ell := |\mathcal{B}_\ell|$ active branches for each node and channel:
\begin{equation}
G^{(\ell)} = \bar X^{(\ell)} W_g + \mathbf{1}{a_g}^\top,
\quad
W_g \in \mathbb{R}^{D \times (B_\ell D)}, a_g\in\mathbb{R}^{B_\ell D}.
\end{equation}
Reshape $G^{(\ell)} \in \mathbb{R}^{N_\ell \times (B_\ell D)}$ into
$\tilde G^{(\ell)} \in \mathbb{R}^{N_\ell \times B_\ell \times D}$ and define
\begin{equation}
\Gamma^{(\ell)}_{i,b,c} =
\frac{\exp(\tilde G^{(\ell)}_{i,b,c})}{\sum_{b'=1}^{B_\ell} \exp(\tilde G^{(\ell)}_{i,b',c})}.
\vspace{-0.1in}
\end{equation}
The fused update is
\begin{equation}
F^{(\ell)} =
\sum_{b=1}^{B_\ell} \Gamma^{(\ell)}_{:,b,:} \odot \Delta^{(\ell,b)},
\qquad
X^{(\ell)} \leftarrow X^{(\ell)} + F^{(\ell)}.
\label{eq:fuse}
\vspace{-0.1in}
\end{equation}
Here, \(\odot\) denotes elementwise multiplication over the node and channel dimensions.

\vspace{-0.1in}
\subsection{Multiresolution encoder and decoder}
\label{sec:encoder_decoder}

We use an encoder and a decoder over the fixed hierarchy. An Adapter maps raw node features to the model width \(D\), producing the finest level encoder features $X_{\mathrm{enc}}^{(0)}=\mathrm{Adapter}(F^{(0)}).$
The encoder applies a SurgFormer block at each level, followed by downsampling:
\begin{equation}
X_{\mathrm{enc}}^{(\ell)} \leftarrow \mathrm{Block}^{\mathrm{enc}}_\ell\!\big(X_{\mathrm{enc}}^{(\ell)}\big),
\qquad
X_{\mathrm{enc}}^{(\ell+1)} \leftarrow P_\ell\!\big(X_{\mathrm{enc}}^{(\ell)}\big),
\qquad \ell=0,\dots,L-1,
\end{equation}
where \(P_\ell\) is the channelwise max pooling operator defined in \eqref{eq:maxpool}. The decoder starts from the coarsest encoder representation, and proceeds from coarse to fine by broadcast upsampling and refinement:
\begin{equation}
\tilde X_{\mathrm{dec}}^{(\ell)} \leftarrow U_\ell\!\big(X_{\mathrm{dec}}^{(\ell+1)}\big),
\qquad
X_{\mathrm{dec}}^{(\ell)} \leftarrow \mathrm{Block}^{\mathrm{dec}}_\ell\!\big(\tilde X_{\mathrm{dec}}^{(\ell)} + X_{\mathrm{enc}}^{(\ell)}\big),
\qquad \ell=L-1,\dots,0,
\end{equation}
where \(U_\ell\) is the broadcast upsampling operator in \eqref{eq:broadcast}. The additive term \(X_{\mathrm{enc}}^{(\ell)}\) provides the same level encoder skip connection.

\vspace{-0.1in}
\subsection{Task definitions, conditioning, and robustness objectives}
\label{sec:tasks_objectives}

\paragraph{Node input features and notation.}
At the finest mesh level with \(N_0\) nodes, each node \(i\) is represented by a base input feature
\(f_i\in\mathbb{R}^{d_{\mathrm{in}}}\), where in our setting \(d_{\mathrm{in}}=7\).
Specifically,
\[
f_i = [p_i \,\|\, s_i \,\|\, c_i],
\qquad
p_i \in \mathbb{R}^3,\;
s_i \in \mathbb{R}^3,\;
c_i \in \{0,1\},
\]
where \(p_i\) is the node coordinate, \(s_i\) is the prescribed tool signal (boundary displacement input), and \(c_i\) is a binary boundary-condition indicator.
Stacking \(\{f_i\}\) yields the raw input matrix \(F^{(0)}\in\mathbb{R}^{N_0\times d_{\mathrm{in}}}\), which is mapped to model width \(D\) by the input Adapter to initialize \(X^{(0)}\).

\paragraph{Prediction head and supervised loss.}
Let \(X_{\mathrm{out}}^{(0)}\in\mathbb{R}^{N_0\times D}\) denote the final decoded finest-level features. For all tasks in this section, we predict nodewise displacements with a shared linear output head
\begin{equation}
\hat U = X_{\mathrm{out}}^{(0)} W_{\mathrm{out}} + \mathbf{1} a_{\mathrm{out}}^\top,
\qquad
W_{\mathrm{out}} \in \mathbb{R}^{D\times 3},\;
a_{\mathrm{out}} \in \mathbb{R}^{3},
\end{equation}
where \(\mathbf{1}\in\mathbb{R}^{N_0}\) is the all-ones vector. Unless otherwise noted, we use mean squared error (MSE) as the supervised loss:
\begin{equation}
\mathcal{L}_{\mathrm{sup}}(\theta;F,U)
=
\frac{1}{N_0}\sum_{i=1}^{N_0}\|f_\theta(F)_i - U_i\|_2^2.
\label{eq:mse_loss}
\vspace{-0.1in}
\end{equation}

\paragraph{Linear FEM deformation.}
For the deformation prediction task, we follow the experimental protocol of Shahbazi et al.~\cite{shahbazi2026neuralaugmentedkelvinletrealtimesoft} and use their single-grasper linear FEM deformation dataset. In this paper, we focus on the network architecture; Kelvinlet-based priors and residual learning are orthogonal to our contributions and are not considered here.

\paragraph{Cut-conditioned deformation.}
For the resection settings described above (e.g., cholecystectomy and appendectomy XFEM simulations), we condition the network on the current cut state using a binary nodewise indicator \(c\in\{0,1\}^{N_0}\), where \(c_i\) encodes whether node \(i\) belongs to the detached/resected side of the interface (equivalently, the procedure-specific cut configuration represented in the XFEM state and detached-vertex flags). We embed this indicator and concatenate it to the base node features before the input Adapter. Specifically, for each node \(i\),
\begin{equation}
e_i = \mathrm{Emb}(c_i),
\qquad
\mathrm{Emb}:\{0,1\}\to\mathbb{R}^{d_e},
\qquad
\tilde f_i = [f_i \,\|\, e_i].
\label{eq:cut_cond}
\end{equation}
Stacking \(\{\tilde f_i\}\) gives \(\tilde F^{(0)}\in\mathbb{R}^{N_0\times(d_{\mathrm{in}}+d_e)}\), which replaces \(F^{(0)}\) as the Adapter input. Thus, ``cut-conditioned deformation'' denotes the same resection-aware prediction setting introduced in the XFEM data-generation section, now represented as an explicit learned conditioning signal. We use the same prediction head and supervised loss in Eq.~\eqref{eq:mse_loss}, with \(F=\tilde F^{(0)}\). We study cross-task transfer from deformation prediction to cut-conditioned deformation in Section~\ref{sec:ablation_transfer}.

\paragraph{Adversarial tool perturbations.} We use adversarial tool signals as a stress test of smoothness and physical plausibility under bounded out-of-distribution, spatially jagged inputs, evaluated on the same linear FEM deformation task as in the deformation modeling. Let \(\mathcal{V}_{\mathrm{surf}}\subseteq \{1,\dots,N_0\}\) be the set of admissible surface nodes. For each adversarial sample, we draw an anchor $\nu \sim \mathrm{Unif}(\mathcal{V}_{\mathrm{surf}})$ and define a small local neighborhood \(\Omega(\nu)=\mathcal{N}_r(\nu)\subseteq \mathcal{V}_{\mathrm{surf}}\).
We parameterize the adversarial tool signal by a local vector \(q_\nu\in\mathbb{R}^3\), and spread it over \(\Omega(\nu)\) using a normalized localized kernel \(\kappa_\nu\in\mathbb{R}^{N_0}\) (zero outside \(\Omega(\nu)\)):
\begin{equation}
S(\nu,q_\nu)=\kappa_\nu q_\nu^\top \in \mathbb{R}^{N_0\times 3}.
\end{equation}
Thus, nodes in \(\Omega(\nu)\) receive a spatially coherent tool signal with shared direction \(q_\nu\). In practice, \(\kappa_\nu\) is a normalized local bump (vMF-style) based on surface-normal alignment. Let \(F^{(0)}=[P\,\|\,S\,\|\,c]\) denote the standard raw input decomposition into coordinates, tool signal, and boundary indicator. For adversarial analysis, we replace the tool channel by \(S(\nu,q_\nu)\), i.e.
\begin{equation}
F^{(0)}_{\mathrm{adv}}(\nu,q_\nu)
=
[P \,\|\, S(\nu,q_\nu) \,\|\, c].
\label{eq:adv_input}
\end{equation}

\noindent Let \(L^{(0)}\) be the finest-level graph Laplacian. For a nodewise displacement field \(U\in\mathbb{R}^{N_0\times 3}\), define the normalized roughness metric
\begin{equation}
\mathcal{M}_{Dr}(U)=
\frac{\operatorname{tr}(U^\top L^{(0)}U)}
{\frac{1}{N_0}\|U\|_F^2+\varepsilon}.
\end{equation}
After standard supervised training, we generate adversarial tool signals by maximizing \(\mathcal{M}_{Dr}\) over \(q_\nu\) while stopping gradients through model parameters:
\begin{equation}
q_\nu^{\mathrm{adv}}(\theta)\in
\arg\max_{\|q_\nu\|_2\le \alpha}
\mathcal{M}_{Dr}\!\big(f_{\operatorname{sg}(\theta)}(F^{(0)}_{\mathrm{adv}}(\nu,q_\nu))\big),
\end{equation}
approximated with projected gradient ascent on \(q_\nu\). We then fine-tune with the clean supervised loss plus an adversarial smoothness penalty on pre-generated adversarial signals \(\{(\nu_m,q_{\nu_m}^{\mathrm{adv}})\}_{m=1}^M\):
\begin{equation}
\min_{\theta}\ 
\mathbb{E}_{(F^{(0)},U)\sim\mathcal D}\!\big[\mathcal{L}_{\mathrm{sup}}(F^{(0)},U;\theta)\big]
+
\lambda_{\mathrm{adv}}\,
\mathbb{E}_{\nu \sim \mathrm{Unif}(\mathcal{V}_{\mathrm{surf}})}\!\big[
\mathcal{M}_{Dr}\!\Big(f_{\theta}\big(F^{(0)}_{\mathrm{adv}}(\nu,q_{\nu}^{\mathrm{adv}})\big)\Big)\big]  
\end{equation}

\vspace{-0.2in}
\section{Experiments}
\label{sec:experiments}

All experiments evaluate nodewise displacement prediction under a shared training and evaluation pipeline, with task-specific conditioning enabled only when applicable. All results are averaged over 3 random seeds; experiments were run on two NVIDIA Blackwell GPUs. The complete training and evaluation configurations are provided in the supplementary material.

\paragraph{Baselines.}
We compare SurgFormer against representative baselines spanning neural-operator and geometric deformation predictors: GAOT~\cite{wen2026geometryawareoperatortransformer}, NIN~\cite{taghikhani2025nin}, MGN-T~\cite{iparraguirre2026mgntransformer}, PointNet~\cite{qi2017pointnetdeeplearningpoint}, and PVCNN~\cite{liu2019pointvoxelcnnefficient3d}. This set covers complementary inductive biases, including operator-style mappings and geometric feature learning.

\paragraph{Metrics and efficiency.}
We report normalized RMSE, normalized Max Err, and DCM~\cite{shahbazi2026neuralaugmentedkelvinletrealtimesoft}; for simplicity, we refer to the normalized error metrics as RMSE and Max Err throughout the paper. DCM measures deformation capture quality relative to the target field. In addition to accuracy, we report inference time per sample and total model size to assess deployment tradeoffs.\\

\noindent To save space, we report mean $\pm$ std only for our primary metric (DCM) and inference time in the main paper, and defer full mean $\pm$ std for RMSE and Max Err to the supplementary material, since their variability closely tracks DCM.

\vspace{-0.1in}
\subsection{Soft-tissue deformation modeling}
\label{sec:soft_tissue}

Table~\ref{tab:all_baselines} reports the baseline comparison for soft-tissue deformation under the protocol as Shahbazi et al.~\cite{shahbazi2026neuralaugmentedkelvinletrealtimesoft}. SurgFormer achieves the best overall accuracy across baselines, with the lowest RMSE (\(0.018\)) and Max Err (\(0.022\)), and the highest DCM (\(97.21\)). It runs at \(0.6\) ms with \(6.5\)M parameters, versus \(0.5\) ms and \(7.2\)M for GAOT, yielding a stronger accuracy-efficiency tradeoff.

\begin{table}[t]
\centering
\scriptsize
\setlength{\tabcolsep}{3.2pt}
\renewcommand{\arraystretch}{1.05}
\caption{Soft tissue deformation modeling results under the same evaluation protocol in \cite{shahbazi2026neuralaugmentedkelvinletrealtimesoft}. SurgFormer achieves the best overall accuracy across baselines, attaining the lowest RMSE and Max Err and the highest deformation capture metric (DCM), while maintaining near real time inference and a comparable parameter budget.}
\label{tab:all_baselines}
\begin{tabular}{@{} l c c c c c @{}}
\toprule
\textbf{Method} &
\textbf{RMSE}$\downarrow$ &
\textbf{Max Err}$\downarrow$ &
\textbf{DCM}$\uparrow$ &
\textbf{Time (ms)}$\downarrow$ &
\textbf{Params (M)} \\
\midrule
GAOT\cite{wen2026geometryawareoperatortransformer}            & 0.028 & 0.034 & 96.36$_{\pm0.88}$ & \textbf{0.53$_{\pm0.04}$} & 7.2 \\
NIN\cite{taghikhani2025nin}             &  0.033   & 0.045 & 93.61$_{\pm0.61}$ & 1.57$_{\pm0.15}$  &  6.8 \\
\midrule
MGN-T\cite{iparraguirre2026mgntransformer}    & 0.083 & 0.122 & 86.76$_{\pm1.27}$ & 1.31$_{\pm0.09}$  & 6.2 \\
PointNet\cite{qi2017pointnetdeeplearningpoint}        &  0.030   &  0.038   & 96.37$_{\pm0.33}$ &  1.28$_{\pm0.05}$  & 6.0 \\
PVCNN\cite{liu2019pointvoxelcnnefficient3d}           &  0.039   &  0.053   & 92.58$_{\pm1.34}$ &  1.69$_{\pm0.23}$  & 6.0 \\
\midrule
\rowcolor{surgblue}
\textbf{SurgFormer} & \textbf{0.018} & \textbf{0.022} & \textbf{97.21$_{\pm1.06}$} & 0.64$_{\pm0.07}$ & 6.5 \\
\bottomrule
\end{tabular}

\vspace{0.6mm}
\footnotesize
\end{table}

\vspace{-0.1in}
\subsection{Cholecystectomy cut-conditioned deformation}
\label{sec:cut}

We evaluate cut-conditioned deformation prediction on the cholecystectomy dataset using the cut-state conditioning interface from Section~\ref{sec:tasks_objectives}. Table~\ref{tab:cut_results} compares models trained with and without cut-state conditioning under the same accuracy and efficiency protocol as Section~\ref{sec:soft_tissue}. Conditioning improves performance across models, indicating that explicit resection-state information is important for post-cut deformation prediction. SurgFormer shows the largest gains, improving RMSE from \(0.143\) to \(0.112\), Max Err from \(0.191\) to \(0.164\), and DCM from \(66.85\) to \(83.61\). It also outperforms GAOT before and after conditioning, with the largest margin in conditioned DCM (\(83.61\) vs. \(72.26\)), while maintaining near real-time inference in a similar range (0.7 ms vs. 0.5 ms).

\begin{table}[t]
\centering
\scriptsize
\setlength{\tabcolsep}{3.8pt}
\renewcommand{\arraystretch}{1.1}
\caption{Cut conditioning results. Each entry reports performance without cut conditioning $\to$ with cut conditioning under the same protocol.}
\label{tab:cut_results}
\begin{tabular}{@{} l c c c c c @{}}
\toprule
\textbf{Method} &
\textbf{RMSE}$\downarrow$ &
\textbf{Max Err}$\downarrow$ &
\textbf{DCM}$\uparrow$ &
\textbf{Time (ms)}$\downarrow$ &
\textbf{Params (M)} \\
\midrule
GAOT\cite{wen2026geometryawareoperatortransformer} &
\makecell{$\text{0.158}\to\text{0.133}$} &
\makecell{$\text{0.289}\to\text{0.221}$} &
\makecell{$\text{63.35}\to\text{72.26}_{\pm1.11}$} &
\textbf{0.51}$_{\pm0.04}$ & 7.2 \\
NIN\cite{taghikhani2025nin} &
\makecell{$\text{0.216}\to\text{0.185}$} &
\makecell{$\text{0.364}\to\text{0.284}$} &
\makecell{$\text{61.11}\to\text{69.78}_{\pm0.93}$} &
1.56$_{\pm0.17}$ & 6.8 \\
PointNet\cite{qi2017pointnetdeeplearningpoint} &
\makecell{$\text{0.199}\to\text{0.157}$} &
\makecell{$\text{0.334}\to\text{0.181}$} &
\makecell{$\text{62.03}\to\text{70.12}_{\pm0.26}$} &
1.33$_{\pm0.06}$ & 6.0 \\
PVCNN\cite{liu2019pointvoxelcnnefficient3d} &
\makecell{$\text{0.227}\to\text{0.144}$} &
\makecell{$\text{0.382}\to\text{0.173}$} &
\makecell{$\text{64.45}\to\text{72.05}_{\pm1.62}$} &
1.59$_{\pm0.18}$ & 6.0 \\
\midrule
\rowcolor{surgblue}
\textbf{SurgFormer} &
\makecell{$\text{0.143}\to\text{\textbf{0.112}}$} &
\makecell{$\text{0.191}\to\text{\textbf{0.164}}$} &
\makecell{$\text{66.85}\to\text{\textbf{83.61}}_{\pm0.44}$} &
0.72$_{\pm0.09}$ & 6.5 \\
\bottomrule
\end{tabular}
\footnotesize

\end{table}

\vspace{-0.1in}
\subsection{Appendectomy procedure}
\label{sec:appendectomy}

Appendectomy includes both uncut deformations and cut-induced deformations under a unified protocol. Each test case provides nodewise inputs and, when applicable, a binary cut indicator. We use the same cut conditioning interface defined in Section~\ref{sec:tasks_objectives}; for uncut cases, we set \(c=\mathbf{0}\). Table~\ref{tab:appendectomy_results} reports averaged results on the mixed test set using the same accuracy and efficiency metrics. In this setting, SurgFormer is within about 1\% DCM of PVCNN, while improving Max Error and delivering about 3× faster inference.

\begin{table}[t]
\centering
\scriptsize
\setlength{\tabcolsep}{3.8pt}
\renewcommand{\arraystretch}{1.1}
\caption{Appendectomy results on mixed uncut and cut cases. Cut conditioning follows \eqref{eq:cut_cond} with $c=\mathbf{0}$ for uncut.}
\label{tab:appendectomy_results}
\begin{tabular}{@{} l c c c c c @{}}
\toprule
\textbf{Method} &
\textbf{RMSE}$\downarrow$ &
\textbf{Max Err}$\downarrow$ &
\textbf{DCM}$\uparrow$ &
\textbf{Time (ms)}$\downarrow$ &
\textbf{Params (M)} \\
\midrule
GAOT\cite{wen2026geometryawareoperatortransformer}        & 0.164 & 0.272 & 79.44$_{\pm1.17}$ & 0.49$_{\pm0.08}$ & 7.2 \\
NIN\cite{taghikhani2025nin}  & 0.155 & 0.263 & 80.24$_{\pm2.24}$ & 0.81$_{\pm0.14}$ & 6.8 \\
PointNet\cite{qi2017pointnetdeeplearningpoint}         & 0.180 & 0.293 & 76.78$_{\pm0.73}$ & 1.07$_{\pm0.13}$ & 6.0 \\
PVCNN\cite{liu2019pointvoxelcnnefficient3d}         & \textbf{0.119} & 0.257 & \textbf{88.74}$_{\pm1.80}$ & 1.23$_{\pm0.19}$ & 6.0 \\
\midrule
\rowcolor{surgblue}
\textbf{SurgFormer} & 0.135 & \textbf{0.228} & 87.61$_{\pm2.02}$ & \textbf{0.48}$_{\pm0.08}$ & 6.0 \\
\bottomrule
\end{tabular}
\end{table}

\subsection{Smoothness under adversarial tool perturbations}
\label{sec:ablation_adv_smoothness}

We use adversarial tool signals (Section~\ref{sec:tasks_objectives}) as a stress test for output smoothness to address the OOD scenarios. Table~\ref{tab:adv_smoothness_stress_test} reports the full evaluation matrix across clean versus adversarial signals and standard versus adversarially fine tuned models. Adversarial signals increase roughness and reduce accuracy for the standard model, while adversarial fine tuning improves smoothness and partially recovers performance under the same perturbation budget. We report one budget \(\alpha\) in the main paper and provide a budget sweep in the supplementary material.

\begin{table}[t]
\centering
\small
\setlength{\tabcolsep}{5pt}
\renewcommand{\arraystretch}{1.08}
\caption{Adversarial smoothness stress test. Clean signals use full supervised metrics. For adversarial tool signals, no target displacement is available, so we report only the smoothness score \(\mathcal{M}_{Dr}\).}
\label{tab:adv_smoothness_stress_test}
\begin{tabular}{@{}llcccc@{}}
\toprule
\textbf{Signal} & \textbf{Model} & RMSE$\downarrow$ & Max Err$\downarrow$ & DCM$\uparrow$ & \(\mathcal{M}_{Dr}\)\(\downarrow\) \\
\midrule
\multirow{2}{*}{Clean}
& Standard & 0.018 & 0.022 & 97.21 & 0.08 \\
& Adv. FT & 0.026 & 0.031 & 95.41 & 0.10 \\
\midrule
\multirow{2}{*}{Adv. (\(\alpha=0.2\))}
& Standard & --- & --- & --- & 0.49 \\
& Adv. FT & --- & --- & --- & 0.11 \\
\bottomrule
\end{tabular}
\end{table}

\vspace{-0.1in}
\section{Ablation Study}
\label{sec:ablation}

We conduct ablations to identify which components of SurgFormer contribute to performance and transfer. We study two factors: the multibranch fusion design and task transfer from deformation prediction to cut conditioned deformation.

\subsection{Design choices}
\label{sec:ablation_design}

\paragraph{Branches, fusion, and learned gate allocation.}
We ablate the multi branch fusion module in \eqref{eq:fuse}, where each level \(\ell\) combines three branch updates from normalized features, \(\Delta_1\) (local), \(\Delta_2\) (global), and \(\Delta_3\) (feedforward), using learned per node and per channel gates \(\Gamma\). We test branch importance by removing one branch at a time and compare learned gating with uniform mixing, \(\Gamma_{i,b,c}=1/|\mathcal{B}_\ell|\) for \(b\in\mathcal{B}_\ell\). As shown in Fig.~\ref{fig:fusion_heatmap_panel} (bottom), the full fusion model performs best (RMSE \(0.018\), Max Err \(0.022\), DCM \(97.21\)). Removing the local branch causes the largest drop (RMSE \(0.026\), Max Err \(0.034\), DCM \(93.84\)), showing that local geometric aggregation is the most important component. Removing the global branch also reduces performance (RMSE \(0.022\), Max Err \(0.029\), DCM \(95.68\)), while removing the feedforward branch has the smallest effect (RMSE \(0.020\), Max Err \(0.025\), DCM \(96.41\)). Learned gating also improves over uniform mixing at the same runtime (\(\sim 0.6\) ms), reducing RMSE from \(0.021\) to \(0.018\) and increasing DCM from \(96.02\) to \(97.21\). This shows that the gain comes from adaptive branch weighting, not only from multi branch capacity. Fig.~\ref{fig:fusion_heatmap_panel} further shows depth dependent gate allocation, supporting learned fusion over fixed averaging.

\newcolumntype{Y}{>{\centering\arraybackslash}X}
\newcolumntype{L}[1]{>{\raggedright\arraybackslash}p{#1}}

\begin{table*}[!th]
\centering
\vspace{-1mm}
\caption{
Multibranch fusion ablation for the fusion operator in \eqref{eq:fuse}. We report design variants obtained by removing local, global, or feedforward branches from the active branch set $\mathcal{B}_\ell$. Branch indices are 1 local, 2 global, and 3 feedforward. Uniform mix uses $\Gamma_{i,b,c}=1/|\mathcal{B}_\ell|$ for $b\in\mathcal{B}_\ell$.}
\label{tab:fusion_ablation}
\scriptsize
\setlength{\tabcolsep}{3.0pt}
\renewcommand{\arraystretch}{1.03}

\begin{tabular}{@{}p{0.31\textwidth}
                S[table-format=1.3]
                S[table-format=1.3]
                S[table-format=2.2]
                S[table-format=1.1]
                S[table-format=1.1]@{}}
\toprule
Variant &
{RMSE$\downarrow$} &
{Max Err$\downarrow$} &
{DCM$\uparrow$} &
{Time (ms)$\downarrow$} &
{Params (M)} \\
\midrule
Full ($\mathcal{B}_\ell=\{1,2,3\}$)        & 0.018 & 0.022 & 97.21 & 0.6 & 6.5 \\
\midrule
No local ($\mathcal{B}_\ell=\{2,3\}$)      & 0.026 & 0.034 & 93.84 & 0.5 & 5.2 \\
No global ($\mathcal{B}_\ell=\{1,3\}$)     & 0.022 & 0.029 & 94.36 & 0.4 & 5.4 \\
No FF ($\mathcal{B}_\ell=\{1,2\}$)         & 0.020 & 0.025 & 96.41 & 0.5 & 6.1 \\
\midrule
Uniform mix ($\mathcal{B}_\ell=\{1,2,3\}$) & 0.021 & 0.027 & 96.02 & 0.6 & 5.9 \\
\bottomrule
\end{tabular}
\end{table*}

\subsection{Task transfer from deformation to cut-conditioned deformation}
\label{sec:ablation_transfer}

\begin{figure*}[!th]
\centering
\vspace{-1mm}

\begin{subfigure}[t]{0.54\textwidth}%
    \vspace{0pt}%
    \centering
    \includegraphics[width=\linewidth]{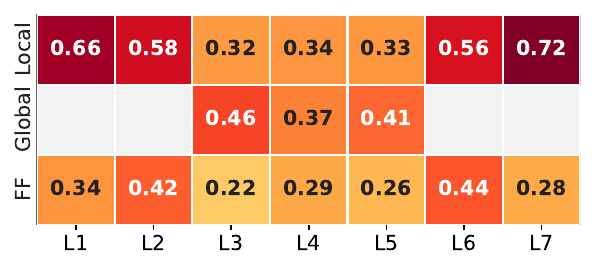}
    \vspace{-0.2in}
    \caption{Fusion gate allocation weights across depth.\vspace{-0.1in}}
    \label{fig:fusion_heatmap_panel}
\end{subfigure}\hfill%
\begin{subfigure}[t]{0.44\textwidth}%
    \vspace{0pt}%
    \centering

    \vspace*{4mm}

    \scriptsize
    \setlength{\tabcolsep}{3.4pt}
    \renewcommand{\arraystretch}{1.08}

    \resizebox{\linewidth}{!}{%
    \begin{tabular}{@{}lccc@{}}
    \toprule
    \textbf{Stage} & \textbf{RMSE}$\downarrow$ & \textbf{Max}$\downarrow$ & \textbf{DCM}$\uparrow$ \\
    \midrule
    Zero shot    & 0.553 & 0.693 & 23.56 \\
    Adapter only & 0.261 & 0.336 & 42.12 \\
    Full FT      & 0.118 & 0.176 & 82.98 \\
    \bottomrule
    \end{tabular}%
    }

    \vspace{-0.5mm}
    \caption{Cross task transfer from deformation modeling to cut conditioned deformation prediction.}
    \label{fig:transfer_table_panel}
\end{subfigure}
\end{figure*}

Table~\ref{fig:transfer_table_panel} evaluates transfer from deformation prediction to cut conditioned deformation using a deformation pretrained checkpoint, with a random cut embedding and reinitialized input weights only where dimensions change. We report zero shot, Adapter warmup (input Adapter + cut embedding only), and full fine tuning. The results show staged adaptation: zero shot performs poorly (RMSE \(0.553\), Max Err \(0.693\), DCM \(23.56\)); Adapter warmup improves substantially (RMSE \(0.261\), Max Err \(0.336\), DCM \(42.12\)); and full fine tuning performs best (RMSE \(0.118\), Max Err \(0.176\), DCM \(82.98\)). This shows that deformation pretraining provides a strong initialization, but accurate cut conditioned prediction requires task specific adaptation beyond the input pathway.

\section{Conclusion}

We present SurgFormer, a multiresolution gated transformer for full field organ deformation prediction on large volumetric meshes with near real time inference. Beyond deformation only surrogates, we study cut conditioned deformation from XFEM generated cutting and resection trajectories within the same volumetric learning pipeline. SurgFormer mixes sparse local message passing at fine scales with coarse global attention, and uses per node, per channel gates to fuse local, global, and pointwise updates, yielding a strong accuracy efficiency tradeoff for interactive surgical simulation. We also introduce two FEM generated datasets for appendectomy manipulation and cholecystectomy resection, and provide unified comparisons across baseline families with robustness analysis to tool signal perturbations via adversarial training and scale invariant Dirichlet regularization. To our knowledge, this is the first learned volumetric surrogate in this setting to evaluate both deformation prediction and cut conditioned deformation under a unified architecture and protocol. Future work includes richer material models, dynamic and history dependent interactions, and tighter coupling between learned surrogates and online simulation control for procedure level training.

\section*{Acknowledgment}
This work was fully supported by the Wellcome-Leap SAVE program.

\bibliographystyle{splncs04}
\bibliography{main}
\clearpage
\newpage

\appendix

\end{document}